\documentclass{article}
\usepackage{arxiv}

\usepackage{amssymb}
\usepackage{amsmath}
\usepackage{color}
\usepackage{enumitem}
\usepackage{amsfonts}
\usepackage{graphicx}
\usepackage{booktabs}
\usepackage{multirow}
\usepackage{tabularx}
\usepackage{float}
  
\usepackage{makecell}
\usepackage{biblatex}
\usepackage{microtype}  

\title{Method and Dataset Entity Mining in Scientific Literature: A CNN + Bi-LSTM Model with Self-attention}

\author{
  Linlin Hou \\
  Center for Combinatorics\\
  NanKai University\\
  Tianjin, China\\
  \texttt{llhou@mail.nankai.edu.cn} \\
   \And
  Ji Zhang\thanks{Corresponding author.}  \\
  Zhejiang Lab\\
  Hangzhou, China\\
  \texttt{zhangji77@gmail.com} \\
  \And
  Ou Wu\\
  Center for Applied Mathematics\\
  Tianjin University\\
  Tianjin, China\\
  \texttt{wuou@tju.edu.cn}\\
  \And
  Ting Yu, Zhen Wang\\
  Zhejiang Lab\\
  Hangzhou, China\\
  \texttt{yuting, wangzhen@zhejianglab.com}\\
  \And
  Zhao Li\\
  Alibaba Group\\
  Hangzhou, China\\
  \texttt{lizhao.lz@alibaba-inc.com}\\
  \And
   Jianliang Gao\\
   Central South University\\
   Changsha, China\\
   \texttt{gaojianliang@csu.edu.cn}\\
  \And
  Yingchun Ye, Rujing Yao\\
  Center for Applied Mathematics\\
  Tianjin University\\
  Tianjin, China\\
  \texttt{yingchunye, rjyao@tju.edu.cn}\\
  }

\date{}
\begin{document}
\maketitle
\begin{abstract}
Literature analysis facilitates researchers to acquire a good understanding of the development of science and technology. The traditional literature analysis focuses largely on the literature metadata such as topics, authors, abstracts, keywords, references, etc., and little attention was paid to the main content of papers. In many scientific domains such as science, computing, engineering, etc., the methods and datasets involved in the scientific papers published in those domains carry important information and are quite useful for domain analysis as well as algorithm and dataset recommendation. In this paper, we propose a novel entity recognition model, called MDER (Method and Dataset Entity Recognition), which is able to effectively extract the method and dataset entities from the main textual content of scientific papers. The model utilizes rule embedding and adopts a parallel structure of CNN and Bi-LSTM with the self-attention mechanism. We evaluate the proposed model on datasets which are constructed from the published papers of four research areas in computer science, i.e., NLP, CV, Data Mining and AI. The experimental results demonstrate that our model performs well in all the four areas and it features a good learning capacity for cross-area learning and recognition. We also conduct experiments to evaluate the effectiveness of different building modules within our model which indicate that the importance of different building modules in collectively contributing to the good entity recognition performance as a whole. The data augmentation experiments on our model demonstrated that data augmentation positively contributes to model training, making our model much more robust in dealing with the scenarios where only small number of training samples are available. We finally apply our model on PAKDD papers published from 2009-2019 to mine insightful results from scientific papers published in a longer time span.
\end{abstract}

\maketitle
\section{Introduction}
Literature analysis plays an important role in contributing to scientific and technological innovation. It facilitates researchers to comprehensively understand the achievements in related fields including the latest advancement and future development trends. How to make it easier and faster for researchers to extract key relevant information and discover useful knowledge from literature papers has started to receive attentions in recent years. \par
With the rapid development of science and technology, the number of scientific papers in literature has grown exponentially in numbers. Traditional literature analysis methods apply manual retrieval and statistical analysis to such a huge amount of papers, which is rather time-consuming and labor-intensive. This has consequently necessitated the application of text mining and related technologies for automated processing in literature analysis.  \par
Literature analysis includes the abstract analysis \cite{Tshitoyan2019Unsupervised}, keywords extraction \cite{Gollapalli2017Incorporating}, the analysis of the cooperation relationship among authors \cite{Liao2018A}, etc. The conventional literature analysis efforts mainly focus on analyzing topics, authors, abstracts, keywords, references, etc., rather than the main content of papers. Yet, we observe that the methods and datasets  involved in the papers in many domains, such as science, computing and engineering, are also important in the scientific literature as they provide readers with important information about the key entities reflecting the methods and datasets involved in the papers. For example, in the sentence of \textit{``As shown later, GRUs gave better results than LSTMs in our settings"}, the method entities are \textit{``GRUs"} and \textit{``LSTMs"}, and in another sentence of \textit{``Finally we report results on CoNLL NER datasets"}, \textit{``CoNLL NER"} here is labeled as a dataset entity. \par
As a relatively new research problem in literature analysis, method and dataset mining in literature can effectively complement the conventional literature analysis and reflect the development trend of methods and datasets and their complex relationships. For example,   scientometrics based on method and dataset mining in scientific literature provide a supplementary to existing techniques based on mining meta data of research papers such as authors and keywords. They can also make more accurate algorithm and/or dataset recommendations if relationships between the existing method and dataset entities are well established. Method and dataset mining from literature is a relatively new research problem, and there has been a relatively small body of reported research on method and dataset mining in scientific papers up till now. Kovacevic et al. \cite{Kovacevic2012Mining} and Houngbo et al. \cite{Houngbo2012Method} adopt a CRF structure to extract entities that contain methods or other semantic entities. Zha et al. \cite{Zha2019Mining} propose a cross-sentence attention network for a comparative relation model to extract the algorithm and algorithm relationship from the text for mining an algorithm roadmap. \par
The primary challenge of method and dataset mining in literature lies in the accurate extraction of the method and dataset entities, where NLP and text mining techniques are the major means to be applied. In recent years, deep neural network methods are become increasingly popular in text mining as they can generate the dense vectors of sentences without handcrafted features, which significantly streamline text learning and analytic tasks \cite{Zhang2017Coupling}. Some research work from different fields have been reported that use deep learning methods to extract data in literature papers from chemistry \cite{Luo2018An}, biology \cite{Li2018Recognizing} and medicine \cite{Ji2018A}. However, these works do not pay much attention to the methods and datasets entities, and their methods cannot directly solve the task in question. Our task involves more fine-grained entity extraction, focusing specifically on method and dataset mining in the scientific literature. \par
In this paper, we propose a novel model, called MDER (Method and Dataset Entity Recognition) for  method and  dataset mining in literature. The model utilizes rule embedding and adopts a parallel structure of CNN and Bi-LSTM with the self-attention mechanism. To facilitate the training and testing of MDER, the datasets are generated from four research areas of computer science, including NLP, CV, Data Mining and AI. By conducting comprehensive experiments on our model, we obtain interesting findings that can provide a good guidance for future practical applications of our model. \par
The main contributions of this paper are summarized as follows: \par
\begin{itemize}
\item We propose MDER, a novel entity recognition framework by incorporating the rule embedding technique and a CNN-BiLSTM-Attention-CRF structure for method and dataset entities mining in scientific literature. MDER is a semantic-based deep learning extraction model which can capture the semantic relationship information between words within sentences.  It incorporates the advantages of multiple components and makes the learning more efficient. Rule embedding reduces the learning burden of the model, CNN and BiLSTM help capture the structural information according to the current context and the self-attention mechanism pays more attention to the important words related to the target entities in sentences.
\item We evaluate MDER on datasets from multiple different research areas of computer science, and the model shows great transfer learning capacity among datasets from different areas. In particular, the model trained on the mixed dataset has the best transferability and generalization performance. Our model also outperforms the state-of-the-art approaches. The ablation experiment demonstrates the effectiveness of each building module of our model in collectively contributing to the good recognition performance of method and dataset entities in literature; \par 
\item Through data augmentation, MDER is capable of effectively dealing with the scenarios where the number of training paper samples is limited, making our model more robust;
\item A long-term literature analysis using MDER to extract entities in PAKDD papers published from 2009 to 2019 shows that our model is effective in analyzing the intrinsic relationships within different methods and datasets exhibiting their development trends over years in a long time span.
\end{itemize}

\section{Related Work}
In this section, we will review related works on literature analysis and named entity recognition.
\subsection{Literature Analysis}
Literature analysis and mining refer to the automatic processing and analysis of the information from a large amount of literature documents. The academic literature, such as the published scientific papers, has unique structured characteristics and is different from other types of literature (news, blogs, web pages, etc.). The academic literature is composed of title, abstract, keywords, main body, and references. The main body usually includes introduction, related work, methods, experiments and conclusions. Early research focuses on mining the bibliographic information (title, authors, references, etc.) of academic literature to study the subject content \cite{Kim2010Semeval, Tan2016Acemap}, such as extraction of keyword from topics, heat analysis of topics, subject classification, author nationality distribution, etc. At the same time, plenty of works have been conducted in abstract keyword analysis, citation relationship analysis, and so on \cite{Gollapalli2017Incorporating, Qazvinian2010Citation, Tan2016Acemap}. Although those approaches seem to work well on summarizing academic papers by investigating bibliographic content, they may miss out a large amount of valuable knowledge hidden in the body of the papers.\par
Knowledge extraction aims to extract various knowledge information (called knowledge elements) by understanding, recognizing, and screening of the knowledge contained in the literature. Knowledge elements are the basic units and structural elements that make up knowledge, and supplement literature metadata (i.e. title, author, abstract, keywords, etc.), and describe the structure of the document. They generally characterize the literature in terms of words, phrases, concepts, and terms, such as research categories, methods, data, indicators, index values, etc. In general, there are four types of knowledge extraction methods: (1) manual annotation-based extraction methods \cite{Augenstein2017Semeval}; (2) pattern-matching based rule extraction methods \cite{Singh2017App}; (3) ontology-based statistical extraction methods \cite{Lin2017Disorder, Okamoto2017Applying}; and (4) deep learning extraction methods \cite{Wagstaff2018Mars, Basaldella2018Bidirectional}. The first three type of methods relying on feature engineering are labor intensive and reflect the coarse semantic granularity of topics and terms. They are applied to construct the domain ontology and reveal the domain development overview, and they cannot provide fine-grained and refined services. Kovacevic et al. \cite{Kovacevic2012Mining} use CRF to identify the four semantics of Task, Method, Resource/Feature and Implementation. Houngbo et al. \cite{Houngbo2012Method} adopt the rule-based technique and CRF to extract method terms from the biomedical corpus, such as algorithm, technique, analysis, approach and method etc. The last category of methods employing deep learning frameworks provides insight of the latent semantic in the literature context. They expand from the literature metadata, topic and term extraction to semantic annotation for natural language, i.e., semantic-based deep learning extraction methods. Zha et al. \cite{Zha2019Mining} propose a cross-sentence attention network for comparative relation model to extract the algorithm and algorithm relationship from scientific publications for mining algorithm roadmap.\par
In this paper, we mainly focus on the methods and datasets appearing in the experimental sections of research publications and use deep learning models to recognize methods and datasets.

\subsection{Named Entity Recognition}
The research work on named entity recognition (NER) \cite{Grishman1996Message}, a fundamental research topic in natural language processing (NLP), has been developed for more than two decades. The NER task aims to recognize naming referential items from the text and classify them into pre-defined and meaningful categories such as names of people, places, and organizations \cite{Yadav2018A}. In specific/different fields, there are various types of defined entities in the corresponding/respective domain. Early studies on NER mainly utilize handcrafted rules with dictionaries and statistical methods. The rule-based methods \cite{Kim2000A,Sekine2004Definition} use the rules built by linguistic experts manually to determine the type according to the matching degree between rules and entities. The statistical-based methods \cite{Kravalova2009Czech} mine and analyze the linguistic information contained in the training corpus by statistical models and extracts features from the training corpus, including word features, context features, and semantic features. These methods are not only difficult to cover all the linguistic scenarios but also time-consuming and labor-intensive.\par
In recent years, deep neural network models are receiving more and more attentions in text mining because they can learn the dense feature representations from raw sequences, instead of using handcrafted features manually engineered by experts \cite{Lample2016Neural, Yang2018Design}. The feature vectors are low-dimensional word representations with rich semantic information. For example, a deep network model, BiLSTM-CRF, is introduced to solve sequence labeling problems for word-level vector representations \cite{Huang2015Bidirectional}. Since then, the BiLSTM structure and the CRF structure have been extensively used in the NER task \cite{Dong2016Character}. In Kim's paper \cite{Kim2016Character-aware}, a convolutional neural network (CNN) was employed over characters to form character representations. Compared with the word-based model, the character-based model \cite{Li2017Leveraging, Yang2017Neural} performs better because the character-based model can handle the input sequence containing unusual characters or out-of-vocabulary words better \cite{Strubell2017Fast, Chen2019GRN}. CNN and BiLSTM are usually used to extract character-level morphological information (such as the prefix or suffix of a word). Ling et al. \cite{Ling2015Finding} build vector representations of words by jointing character representations which are obtained through using BiLSTM. Chiu et al. \cite{Chiu2016Named} and Ma et al. \cite{Ma2016End} concatenate word embedding and character embedding which are obtained through a CNN structure as the input vector representations of the model. Moreover, using the attention mechanism can enhance the word representation for concentrating on the key part of a sentence \cite{Zukovgregoric2017Neural, Luo2018An}. However, the models proposed by these works are relatively simple which only use either the embedding of characters and words, or the attention mechanism. Usually, these models structurally connects CNN and LSTM in series which might lose some information in the transmission process. Also, they are not very effective in capturing the connections between words and the dependencies between tags. In our paper, CNN and LSTM are used in parallel which makes the output expression vector retain more information. Our multiple components are integrated by adding self-attention and CRF which could better characterize the connection between different words and the dependency between tags. \par

There have been also a few surveys which focus on NER systems for specific domains and languages, including biomedical NER \cite{Gridach2017Character, Wei2019Named}, Chinese clinical NER  \cite{Wang2019Incorporating, Li2019An} and chemical NER \cite{Korvigo2018Putting}.

\section{Our Model}
In this section, we will elaborate on the model we propose for recognizing methods and datasets from scientific literature. We start this section by first introducing the necessary notational conventions and the formulation of our research problem as follows.\par

\begin{figure}
	\centering
		\includegraphics[scale=.64]{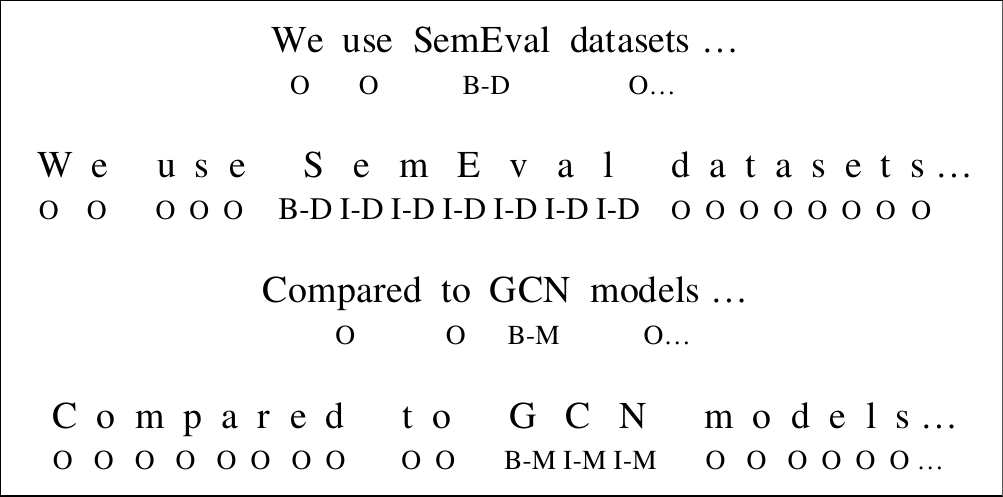}
	\caption{An example sentence with word level and character level NER tags.}
	\label{Figure: 1 example}
\end{figure}

Given a sentence $\{w_1, w_2, \cdots, w_n\}$ can define its character-level sequence as $\{c_{11},c_{12},\cdots,c_{ij},\cdots$ $,c_{nm_n}\}$, where $w_i$ represents the $i$-th word, $n$ is the length of the sentence, $c_{ij}$ denotes the $j$-th character of the $i$-th word and $c_{nm_n}$ denotes the $m_n$-th character of the $n$-th word. For the sake of simplicity, we will rewrite the character-level input sequence of the given sentence as $\{x_1,x_2,\cdots,x_i,\cdots,x_m\}$, where $x_i$ is the $i$-th character of the sequence with a length of $m$. Our aim is to identify the tags (e.g. $\emph{M}$, $\emph{D}$, or others) of all characters for each word \cite{Dong2016Character, Akbik2018Contextual}. To perform the task, we construct an entity recognition classifier that recognizes the entities in the sentence. We note that NER is first proposed as a word-level tagging problem, and most existing datasets use word-level tags to denote named entity phrases. However, as mentioned before, we treat a sentence as a sequence of characters in this paper and our model is designed to predict the tag of each character. Therefore, we choose to use character-level tags. 
As a major advantage, character-level tagging can, to a large extent, avoid the appearance of new words when predicting character tags. We apply the BIO (Begin, Inside, Outside) tagging scheme for characters. Let $L=\{B-M, I-M,B-D,I-D,O,padding\}$ denote the tagging set, “B–M” and “B-D” denote the tags of the beginning character of the method entity and the dataset entity, respectively, “I-M” and “I-D” represent the tags of the inside character of the method entity and the dataset entity, respectively. If a character subsequence in a sentence constitutes a named entity (NE) phrase, each character in that subsequence receive the tag composed of the position indicator (B, I) and the NE type (M, D). Otherwise, characters are assigned with the outside tags (O). For example, the tag of the method entity, “LSTM”, is “B-M, I-M, I-M, I-M”. The goal of our work is to essentially map each sentence into a sequence of character-level tags. As a comparison, Figure \ref{Figure: 1 example} shows the difference between the word-level and character-level tags for an example sentence.\par

\begin{figure}
	\centering
		\includegraphics[scale=.25]{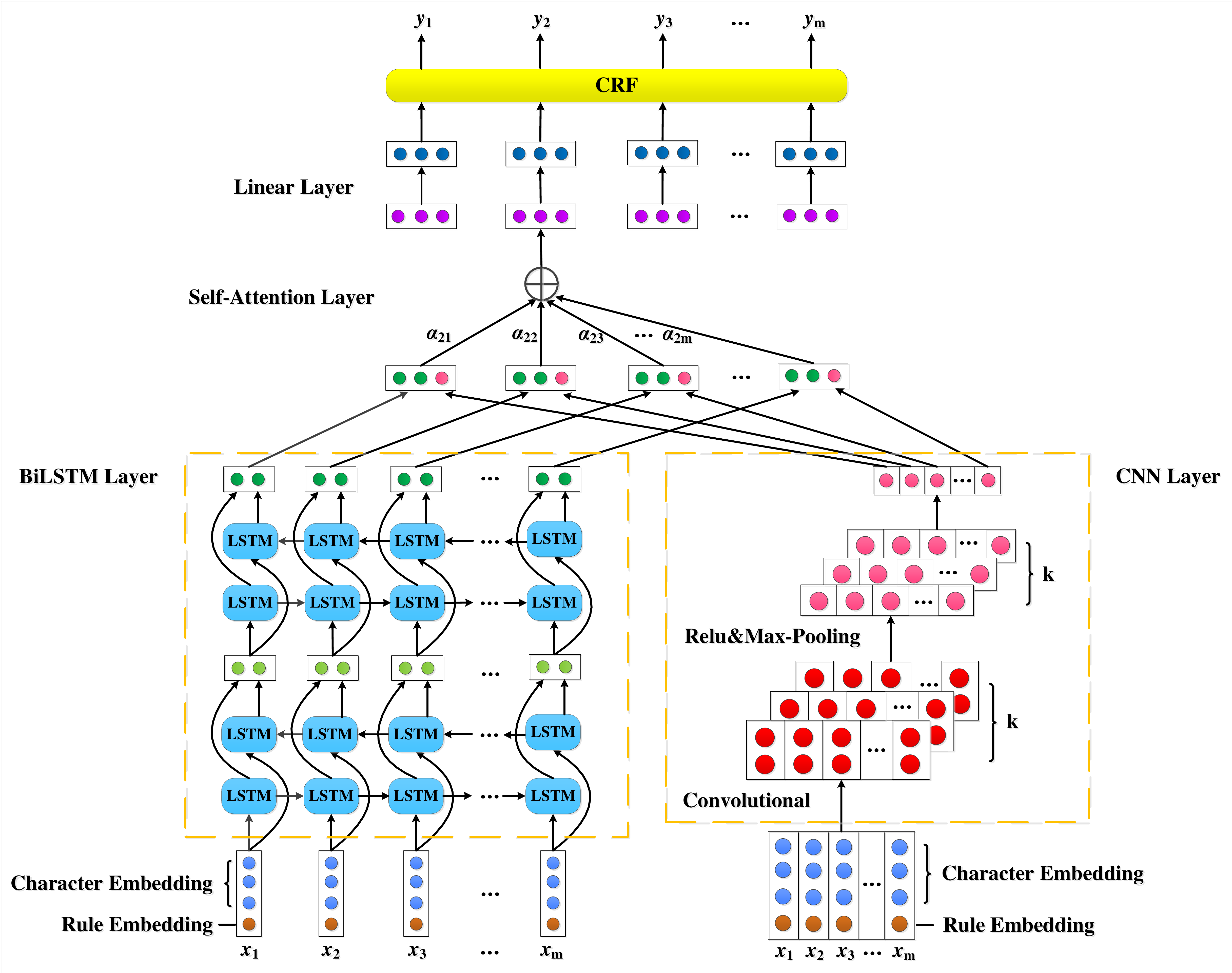}
	\caption{The architecture of our MDER model.}
	\label{Figure:2 MDER model}
\end{figure}

Our model combines the rule-based technique and a new deep neural network structure. The overall architecture of the proposed model is illustrated in Figure \ref{Figure:2 MDER model}. It consists of the input embedding layer, the BiLSTM-CNN layer, the self-attention mechanism layer and the CRF output layer.  Rule embedding can reduce the learning burden of the model and make the learning more efficient. CNN is a useful supplement to BiLSTM which helps capture the structural information according to the current context. The self-attention mechanism pays more attention to the important words related to the target entities in sentences and interactive information between different words. CRF considers the correlations between tags in neighborhoods and decodes the best tag chain jointly. Our model incorporates the advantages of multiple components and makes the learning more effective. Next, we will explain all components in the model sequentially from the input to the output.

\subsection{Input Embedding Layer}

The input embedding layer contains both the rule embedding and character embedding. The rule embedding is based on the concept of user-specified blacklist and whitelist which reflect the regular pattern of special entities in the specific scientific domain in question, while the character embedding directly transforms the words in literature by mapping each character to a high-dimensional vector space.

\subsubsection{Rule Embedding}
We construct one blacklist containing some general words and two whitelists of entities. The tag of each character in the blacklist and whitelists are regarded as additional supervised information which helps train the model parameters. The whitelist of methods contains some common method entities, such as “SVM”, and the tags of characters in this word are B-M, I-M and I-M, respectively. The whitelist of datasets contains some known dataset entities, such as “Wiki” whose tagging sequence is \{B-D, I-D, I-D, I-D\}. The blacklist contains some general words such as “the” with character tags \{O, O, O\}. The characters of the words that do not belong to the blacklist or whitelists are set to unknown. \par
For rule embedding, each character adopts the aforementioned blacklist and whitelist tagging method. Then, each $x_i$ is represented using $x_i^r=e^r(x_i)$, where  $e^r\in \mathbb{R}^{d_{r\times m}}$ denotes a rule embedding lookup table matrix obtained through model learning, and $x_i^r\in \mathbb{R}^{d_{r\times 1}}$. $d_r$ is the dimension of rule embedding of each character.

\subsubsection{Character Embedding}

For character $x_i$, its character embedding is represented as $x_i^c:=e^c(x_i)$ by searching the  character embedding lookup table matrix $e^r\in \mathbb{R}^{d_{c\times m}}$, where $x_i^c\in \mathbb{R}^{d_c\times 1}$, $d_c$ is the dimension of character embedding of each charter.\par
The final feature representation of a character is the concatenation of its character embedding and its rule embedding as $x'_i=[x_i^r; x_i^c]$. For an input sentence $[x'_1, x'_2, \cdots, x'_m]$, each character representation is an embedding vector $x'_i\in \mathbb{R}^{(d_r+d_c)\times 1}$, where $d_r+d_c$ is the dimension of the final character vectors. \par

\subsection{BiLSTM-CNN Layer}

In the subsequent embedding layer, character representations $[x'_1, x'_2, \cdots, x'_m]$ are fed into both a BiLSTM structure and a CNN structure. For the BiLSTM-CNN layer, the input is the character sequence of one sentence as $\{x_1,x_2, \cdots, x_i,$ $\ldots, x_m\}$.

\subsubsection{Bidirectional Long Short-Term Memory (BiLSTM)}

Bidirectional Long Short-Term Memory (BiLSTM) \cite{Graves2005Framewise} has been shown powerfully to capture the dependencies of the input sequence. We employ a two-layer BiLSTM on top of the embedding layer. The output of the first layer can capture more syntactic information, while that of the second layer can learn more semantic information. \par
In our model, the forward pass of a unidirectional LSTM \cite{Jozefowicz2015An, Hochreiter1997Long} at the $t$-th input character is calculated as follows: \par
\begin{align}\label{1-6}
i_t&=\sigma(W_i[h_{t-1}; x_t]+b_i) \\
f_t&=\sigma(W_f[h_{t-1}; x_t]+b_f) \\
o_t&=\sigma(W_o[h_{t-1}; x_t]+b_o) \\
\tilde{c_t}&=\tanh (W_c[h_{t-1}; x_t]+b_c) \\
c_t&=f_t*c_{t-1}+i_t*(\tilde{c_t}) \\
h_t&=o_t*\tanh (c_t)
\end{align}
where $W_i$, $W_f$, $W_o$ and $W_c$ are the parameter matrices to be learned, $b_i$, $b_f$, $b_o$ and $b_c$ are the bias parameters, $\sigma$ is the logistic sigmoid active function, tanh is the hyperbolic tangent function, $\ast$ is element wise multiplication, $x_t$ is the input of the current time $t$, $h_{t-1}$ is the output of the hidden layer at timestamp $t-1$ and $h_t$ is the output of the hidden layer at the current time $t$. $i_t$, $f_t$ and $o_t$ are the input gate, forget gate and output gate,  respectively, $c_t$ and $c_{t-1}$ are the current cell state and the state of the cell at the previous moment, respectively and $\widetilde{c_t}$ is temporary cell state at the current time $t$. \par
After feeding character representation to the first layer of the two-layer BiLSTM, the BiLSTM units will generate the forward hidden states $\{\overrightarrow{h_1^1}, \ldots, \overrightarrow{h_m^1}\}$ and the backward hidden states $\{\overleftarrow{h_1^1}, \ldots,\overleftarrow{h_m^1}\}$, where $\overrightarrow{h_t^1}, \overleftarrow{h_t^1} \in \mathbb{R}^{d_h\times1}$.  By concatenating the two hidden states, the first layer of BiLSTM outputs the intermediate result $h_t^1=[\overrightarrow{h_t^1};\overleftarrow{h_t^1}]$. $h_t^1$ is also the input of the second layer of BiLSTM at timestamp $t$. ${h_1^1, \ldots,h_m^1}$ go through the same operation as above to produce the final output of the second layer of BiLSTM. Finally, the forward state $\overrightarrow{h_t^2}$ and the backward hidden state $\overleftarrow{h_t^2}$ are concatenated as the final output representation of the two BiLSTM layer as: \par

\begin{align}\label{7,8}
h_t&=[\overrightarrow{h_t^2};\overleftarrow{h_t^2}] \\ 
H&=\{h_1, \ldots,h_m\}
\end{align}
where $h_t\in \mathbb{R}^{2d_h\times 1}, H_t\in \mathbb{R}^{2d_h\times m}$ and $d_h$ denotes the number of hidden units.

\subsubsection{Convolutional Neural Network (CNN)}

Previous works have shown that Convolutional Neural Network (CNN) is an effective approach to extract character-level morphological information from characters of words \cite{Santos2014Learning, Chiu2016Named}. Therefore, we use CNN in our work to capture this structural information according to the current context and encode characters into neural representations. 
We use  \emph{k} convolution kernels with size $p\times q$ and the convolution stride $s\times t$ to execute a convolution operation on inputs $[x'_1, x'_2, \ldots, x'_m]$ and obtain $k$ feature maps. Then, the feature maps are operated by the Rectified Linear Unit (ReLU) activation function. The ReLU function is computed as follows: \par
\begin{align}\label{9}
relu(e)=
\begin{cases}
0, & \text{if $e\leq 0$}\\
e, & \text{if $e> 0$}
\end{cases}
\end{align}
where $e$ is an element of one feature map. Each element of one feature map is fed into Formula \ref{9} to produce a new feature map. Then, by applying the maximum pooling to new feature maps $M^i$, we obtain the flattened representation of each new feature maps as
\begin{align}\label{10}
c^i={MaxPooling}_{t=1,\cdots,m}(M^i)
\end{align}
where $c^i\in \mathbb{R}^m, i=1,2,\cdots, k$. \par

Finally, we stack $k$ feature vectors together. The output vector $h'_t \in \mathbb{R}^{d_{cnn}\times 1}$ of CNN corresponding to each character is obtained by concatenating the same position of feature representations $\{c^1, c^2,\cdots, c^k\}$. In this way, we obtain the final representations $\{h'_1, h'_2,\cdots, h'_m \}$ of CNN layer. Note that here $d_{cnn}$ is equal to $k$. \par
The output vector $h_t$ of BiLSTM and the output vector $h'_t$ of CNN  are concatenated as the input of the next layer as
\begin{align}\label{11, 12}
g_t&=[h_t;h'_t]\\
G&=\{g_1,\cdots,g_m\}
\end{align}
where $g_t\in \mathbb{R}^{(2d_h+d_{cnn})\times 1}$ and $G\in \mathbb{R}^{(2d_h+d_{cnn})\times m}$.

\subsection{Self-attention Layer}

To capture the interactive information between the context and the character, we employ a self-attention mechanism \cite{Vaswani2017Attention, Tan2018Deep} in our model. The self-attention mechanism can capture the long-range dependencies between tokens and contextual information in a sequence. It selectively pays more attention to some important information and gives higher weights to them, while gives lower weights to other information. Now, we describe how we obtain the attention vector representation. We define that:\par
\begin{align}\label{13-15}
Q&=G^TW^Q\\
K&=G^TW^K\\
V&=G^T
\end{align}
where $W^Q, W^K\in \mathbb{R}^{(2d_h+d_{cnn})*d_Q}$ are parameters to be learned during the training. For each character representation in the sequence, we take the dot product of its linear transformation vector and linear transformation vector of every character representation in the sequence to calculate un-normalized attention weights. We adopt the softmax of the weights to produce the normalized attention weights matrix $\boldsymbol{\alpha}$, where $\boldsymbol{\alpha}\in \mathbb{R}^{m\times m}$ and softmax(·) is a column-wise normalizing function. The standard self-attention mechanism is computed as follows: \par
\begin{align}\label{16}
\boldsymbol{\alpha}=softmax(QK^T)
\end{align}\par
Then, we use the attention weights $\boldsymbol{\alpha}$ to create a weighted sum across all output vector of the BiLSTM-CNN layers for attention vector representation $h_i^a$:
\begin{align}\label{17-19}
&H^a=Attention(Q,K,V)=\boldsymbol{\alpha}V\\
&i.e.\ h_i^a=\sum_{j=1}^{m}{\alpha_{ij}V_j}\\
&\sum_{j=1}^{m}\alpha_{ij}=1, \forall i=1, 2, \ldots, m
\end{align}
where $h_i^a$ denotes the component of $H^a$, and $h_i^a\in \mathbb{R}^{1\times (2d_h+d_{cnn})}$. As an attention vector at position $i$, $\alpha_{ij}$ indicates how much attention received at position $j$. $H^a=(h_1^a, \cdots, h_m^a)$ is the attention vector representations which each captures the history information of the whole sentence.

\subsection{Linear Layer}
Next, a linear layer is added on top of the attention layer and is used to produce the probability score for each character. Hence, the final output representation of the self-attention layer is computed by a full connection which maps $h_i^a$ into the tagging space of $|L|$ classes:
\begin{align}\label{20}
Z=H^aW^a+b^a
\end{align}
where$Z,b^a\in \mathbb{R}^{m\times |L|}, W^a\in \mathbb{R}^{(2d_h+d_{cnn})\times |L|}$. $b^a$ is the bias vector, $W^a$ is the transformation matrices and $|L|$ is the size of tagging set ($|L|=6$).

\subsection{CRF Output Layer}

A conditional random field (CRF) \cite{Lafferty2001Conditional} is a random field globally conditioned on the observation sequence and has been widely used in feature-based supervised learning approaches. Many deep learning based NER models use a CRF layer as the tag decoder because of its ability to consider the correlations between tags in neighborhoods and decode the best tag chain jointly \cite{Zheng2017Joint, Strubell2017Fast}. So, we also use a CRF module to jointly decode tag sequences to extract entities. We consider $Z$ to be the input sequence scores, which is generated from the self-attention layer. Therefore, for an input sentence (character-level) $x=\{x_1,x_2,\cdots,x_m\}$, the probability score of the $i$-th character being assigned with the $j$-th tag is calculated as $Z_{ij}$, where $i=1,2,\cdots,m$ and $j=1,2,\cdots,|L|$. We define the sequence of expectations as $y=\{y_1,y_2,\cdots,y_m\}$, where $y_i\in \{B-M,I-M,B-D,I-D,O,padding\}$. Then, we set $A$ as the transition matrix of probability $Z$, and $A_{ls}$ is the transition probability from tag $l$ to tag $s$ , where $l, s\in L$. The decoding score of prediction $y=\{y_1,y_2,\cdots, y_m\}$ is computed as:
\begin{align}\label{21}
score(x,y)=\sum_{i=1}^{m}{Z_{i,y_i}+\sum_{i=0}^{m}A_{y_i,y_{i+1}}}
\end{align}
where $Z_{i,y_i}$ is the score of tag $y_i$ for character $x_i$, and $A_{y_i,y_{i+1}}$ corresponds to the transition score of tag $y_i$ to tag $y_{i+1}$. For each $y$, we use a softmax function to calculate the conditional probability over all possible tag sequences as follows: \par
\begin{align}\label{22}
p(y|x)=\frac{exp(score(x,y))}{\sum_{\tilde{y}\in Y_x}exp(score(x,\tilde{y}))}
\end{align}
where $Y_x$ is the set of the possible tag sequences for given $x$. \par

During the training stage, we train the model parameters by maximizing the log-likelihood probability of correct tag sequence on the training set $\{(x,y^x)\}$, where $y^x$ is the true tag sequence for input sentence $x$. The calculation formula is defined as follows: \par
\begin{align}\label{23}
ln(p(y^x|x))=score(x,y^x)-log(\sum_{\tilde{y}\in Y_x}exp(score(x,\tilde{y})))
\end{align}\par
Finally, during the tag sequence prediction, we derive the optimal $y^*$ such that Equation \ref{24} can be maximized and take it as the predicted tag sequence, as shown in the following equation: \par
\begin{align}\label{24}
y^*={argmax}_{\tilde{y}\in Y_x}score(x,\tilde{y})
\end{align} \par
Consequently, we obtain the tagging sequence $y^*$ with the highest score among all tagging sequences in the final decoding stage using the dynamic programming Viterbi algorithm which is often used to solve the optimal path problem.

\section{Experimental Evaluation}
In this section, we experimentally evaluate the proposed model under different datasets against multiple evaluation metrics. 

\subsection{Experimental Data Construction}

Existing entity recognition datasets, either available  publicly Or used by other existing works, are more about the recognition of proper nouns such as places, organizations, people names, chemical or medical entities, etc., which are not suitable for training and evaluating our model. Hence, we need to construct new datasets to cater for the special needs of this research work. In order to study the practicability of the proposed model in recognizing method and dataset entities in scientific literature of different domains, we first construct four new datasets in this research based on the published papers from top-tier flagship conferences in four popular research areas of computer science, namely NLP, computer vision, data mining and artificial intelligence. They are respectively the Annual Meeting of the Association for Computational Linguistics (ACL), the International Conference on Computer Vision and Pattern Recognition (CVPR) , the ACM International Conference on Knowledge Discovery and Data Mining (SIGKDD) and AAAI Conference on Artificial Intelligence (AAAI). We downloaded 50 papers from the official websites of ACL \footnote{https://www.aclweb.org/anthology/events/acl-2019/}, CVPR \footnote{http://openaccess.thecvf.com/CVPR2019.py}, SIGKDD \footnote{https://www.kdd.org/kdd2019/accepted-papers\#} and AAAI \footnote{https://www.aaai.org/Library/AAAI/aaai19contents.php}, respectively. \par 
The construction of each of the four area-specific datasets takes three steps: 1) segmentation of the sentences in the experimental section of the papers, 2) sentence preprocessing and 3) manually tagging entities. Firstly, we convert the PDF version of a paper into its TXT text version, use rule matching to extract the entire paragraphs of the experimental section of the paper and cut paragraphs into sentences by punctuation (e.g. \textit{``."}, \textit{``?"}, etc.). Secondly, we correct the spelling mistakes that are generated during the paper format transformation process. The same number of sentences are randomly selected from all the four areas to generate the four datasets. Because the minimum number of sentences collected in the four area is 2,800, so the number of sentences for the four datasets are all kept as 2,800 to facilitate the experiment. \par

Thirdly, we recruit six graduate students in our institute to manually tag these sentences, which followed a standard process. Every pair of students marked the same sentences, and the tags are considered to be correct if both of them have the same tagging result. As a result, each dataset ends up with 2,800 annotated sentences. The details of the four datasets are shown in Table \ref{Table: 1 datasets}. \par

\begin{table}[htbp]
    \centering
    \small
    \setlength{\tabcolsep}{4pt}
    \begin{tabular}{lccccc}
    \toprule
    Data&\#sentence&\#sentence&\#method&\#dataset&\#entities \\
        &          &with entities&entities&entities&  \\
    \midrule
    \multirow{1}{*}{ACL}     &2,800&1,322&1,725&626&2,351 \\
    \cmidrule(lr){1-6}
    \multirow{1}{*}{CVPR}    &2,800&1,301&1,403&827&2,230 \\
    \cmidrule(lr){1-6}
    \multirow{1}{*}{SIGKDD}  &2,800&1,376&2,017&497&2,514 \\
    \cmidrule(lr){1-6}
    \multirow{1}{*}{AAAI}    &2,800&1,453&2,072&611&2,683 \\
    \bottomrule
    \end{tabular}
\caption{\label{Table: 1 datasets} The details of the experimental data sets.}
\end{table}

In addition to the four area-specific datasets, we also construct a mixed dataset based on the four datasets. The mixed dataset is constructed via randomly selecting 700 sentences from each of the four datasets ACL, CVPR, SIGKDD and AAAI, then combining them to form the 2,800 sentences. \par
All the five datasets help us better study the cross-area transferability of our model, while at the same time ensuring the consistency of the experiments when evaluating our model in different areas.

\subsection{Experimental Settings}
Each data corpus is randomly divided into two parts with the proportion of 4:1, with four folds are used for training and one fold is used for testing. One eighth of samples in training data are used as the validation data. In other words, each dataset is split into training, validation and testing sets with the ratio of 7: 1: 2. Since each dataset has 2,800 annotated sentences, so the number of the corresponding training set, cross-validation set and test set is 1960, 280 and 560, respectively. \par

In our experiments, we set reasonable values for various parameters used. For each input sentence, the maximum length of its character sequence is set to 600, the dimension of the rule embedding is 40 and the character embedding dimension is 200. To reduce feature dimension, save the computational overhead and enhance the nonlinear expressiveness of the model, the CNN component of the proposed model uses 30 convolution kernels with size $1\times 1$ and the convolution stride is $1\times 2$ (We tested other choices as well, but this set of values is the best). Thus, $d_{cnn}$ and $k$ are both 30. BiLSTM module contains two layers, with each layer having 200 hidden units. In the attention layer, the second dimension $d_Q$ of the involved matrix $W^Q$ is 400. During the training stage, the batch size is set to 16 and we minimize the loss function using the Adam optimizer \cite{Kingma2014Adam} in MDER. The dropout rate \cite{Srivastava2014Dropout} and learning ratio are set to 0.5 and 0.001, respectively. Our model is implemented using Tensorflow. \footnote{https://www.tensorflow.org/} \par

To evaluate the performance of MDER, we employ three types of evaluation metrics, Precision (P), Recall (R) and F1-score (F1). The F1 is the harmonic mean of P and R, which can provide a more balanced evaluation of the performance of the model. Specifically, Precision, Recall and F1-score are defined as follows: \par
\begin{align}\label{24-25}
&Precision=\frac{\#\ correct\ entities}{\# identified\ entities} \\
&Recall=\frac{\#\ correct\ entities}{\#\ annotated\ entities} \\
&F1=\frac{2\times Precision\times Recall}{Precision+Recall}
\end{align}
where $\#\ identified\ entities$ denotes the number of entities that are predicted, $\#\ correct\ entities$ represents the number of entities that are correctly predicted, and $\#\ annotated\ entities$ denotes the number of entities that are in the dataset corpus. Note, an entity is considered to be predicted correctly only if the label of every character of the entity is correctly predicted.

\subsection{Effectiveness Evaluation and Analysis}

\begin{figure}[H] 
	\centering \includegraphics[scale=0.38]{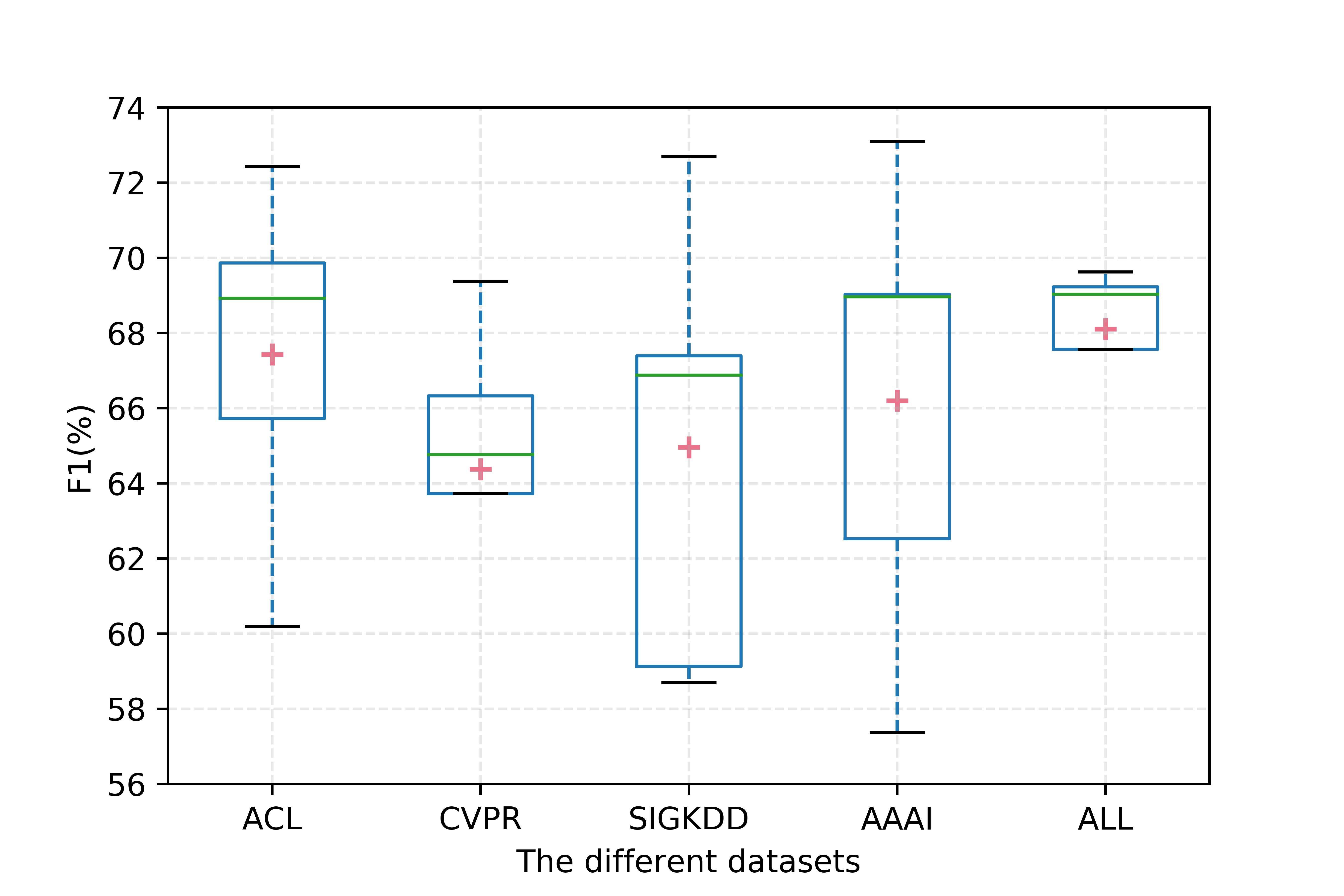}
	\caption{The Box-plot of F1 values of each model.}
	\label{Figure: 3 Box-plot}
\end{figure}

Using F1 measure, we study the effectiveness performance of MDER which is trained using the training samples and tested using the test samples of the five different datasets. This experiment helps the evaluation of the transferring learning performance of MDER, which refers to the performance of applying MDER, which is trained using the data in one area, to the data of another one. For MDER, we conduct the experiment three times on each dataset and report the average test results. The mixed dataset samples multiple times from the other four datasets to repeatedly train the model. Table \ref{Table: 2 Experimental result} shows the F1-meansure experimental results of our models trained on different datasets. The result of the $i$-th row and the $j$-th column in the table indicates that MDER uses the $i$-th dataset as the training set and the validation set, and uses $j$-th dataset as the test set. For example, the value of 0.6893 in the first row and the fourth column of the table means that the model trained with the ACL dataset achieves F1 score of 0.6893 on the testing samples of the AAAI dataset. Here, 70\% of the ACL dataset is used as the training set, 10\% of the ACL dataset is used as the verification set, and 20\% of the AAAI dataset is used as the test set. In this way, we obtain a total of 25 test results as shown in Table \ref{Table: 2 Experimental result}. We also calculate the average and standard deviation of the test results of each model and present them in the last two columns of Table \ref{Table: 2 Experimental result}. Furthermore, Figure \ref{Figure: 3 Box-plot} depicts F1 values of each model on the five test sets in a boxplot.\par

\begin{table}[!htbp]
\centering
\setlength{\tabcolsep}{4pt}
\begin{tabular}{|l|l|c|c|c|c|c|c|c|}
\hline
\multicolumn{2}{|c|}{\multirow{2}*{F1}}& \multicolumn{5}{c|}{Test Data} &\multicolumn{2}{c|}{Statistics}\\
\cline{3-9}
\multicolumn{2}{|c|}{}&ACL&CVPR&SIGKDD&AAAI&Mixed&mean&std\\
\hline
\multirow{4}*{\makecell[c]{Traning\\Data}}&ACL&0.7243&0.6020&0.6573&0.6893&0.6987&0.6743&0.0420\\
\cline{2-9}
&CVPR&0.5770&0.6937&0.6373&0.6477&0.6633&0.6438&0.0384\\
\cline{2-9}
&SIGKDD&0.5870&0.5913&0.7270&0.6688&0.6740&0.6496&0.0534\\
\cline{2-9}
&AAAI&0.6253&0.5737&0.6897&0.7310&0.6903&0.6620&0.0556\\
\cline{2-9}
&Mixed&0.6507&0.6757&0.6903&0.6923&0.6963&0.6811&0.0167\\
\hline
\end{tabular}
\caption{\label{Table: 2 Experimental result} The results of our model on different train/test datasets.}
\end{table}

From the results presented in Table \ref{Table: 2 Experimental result} and Figure \ref{Figure: 3 Box-plot}, we can obtain a rich set of interesting findings as follows: \par
1) We can see from Table \ref{Table: 2 Experimental result} that the diagonal cells within the table have the highest F1 scores for each row of the result matrix. It means the model achieves the best predicting results when the training and test sets are from the same area. This result is intuitive and easy to understand since both sets are sampled from the same domain and therefore share very similar features in terms of entity types, syntactic and semantic information; \par
2) For each row in Table \ref{Table: 2 Experimental result}, the second overall highest F1 value appears in the column of the mixed dataset. This is because one quarter of the training set of the dataset come from each of the other four area-specific datasets, whereby providing a fair amount of area-specific information for training the model; \par 
3) The third overall highest F1 value of each row is in the column of AAAI dataset because AAAI covers a broad sub-fields of artificial intelligence including NLP, computer vision and data mining, even though it is not as representative as the mixed dataset we constructed; \par
4) The model trained by the mixed dataset has the highest average F1 value and the smallest variance when compared with the models trained using the other four datasets. This suggests that the model trained by the mixed dataset features the best entity recognition and generalization performance across different areas under study. The model trained using the other four area-specific datasets tend to be inferior in terms of performance due to the more limited area-specific entity features learned from the training sets. This also demonstrates the salient advantage of creating the mixed dataset in contributing to a more robust model that excels in transferring learning when mining methods and datasets from literature across multiple different areas.

\subsection{Comparative Study}

\begin{table}[htbp]
    \centering
    \small
    \setlength{\tabcolsep}{4pt}
    \begin{tabular}{lccccccc}
    \toprule
    &Precision&Recall&F1 \\
    \midrule
    \multirow{1}{*}{Baseline1}     &0.660&0.562&0.6071 \\
    \cmidrule(lr){1-4}
    \multirow{1}{*}{Baseline2}     &0.665&0.569&0.6133 \\
    \cmidrule(lr){1-4}
    \multirow{1}{*}{Baseline3}     &0.698&0.597&0.6436 \\
    \cmidrule(lr){1-4}
    \multirow{1}{*}{MDER}     &$\textbf{0.7623}$&$\textbf{0.6420}$&$\textbf{0.6963}$ \\
    \bottomrule
    \end{tabular}
\caption{\label{Table: 3 Comparative result}  Our model and other model on the mixed dataset.}
\end{table}

As far as we know, there are very limited research works on the problem of method and dataset mining from literature. To evaluate the performance of our proposed model through a comparative study, we select several popular models for named-entity recognition (NER) as the baseline methods for performance comparison, which include:\par
\begin{itemize}
\item Kuru et al. \cite{Kuru2016Charner} use a BiLSTM + softmax network for CharNER (called Baseline1);
\item Huang et al. \cite{Huang2015Bidirectional} use a BiLSTM + CRF model for sequence tagging. The BiLSTM + CRF structure is widely used as a baseline model for the sequence labeling work (called Baseline2);
\item Gregoric et al. \cite{Zukovgregoric2017Neural} construct a BiLSTM + Self-Attention + CRF network for NER (called Baseline3).
\end{itemize}
Our model and the three baseline models are trained and tested three times on the mixed dataset and results are averaged. To evaluate the performance of models, we employ precision, recall, and F1-score metrics. Table 3 shows the experimental results of the comparative study. As depicted in the last row of Table \ref{Table: 3 Comparative result}, our model offers a significantly better performance in all the three measurements compared with the three baseline models.

\subsection{Ablation Study (The Effect of Different Modules)}

Since our proposed model consists of several key building modules (i.e., rule embedding, CNN (two-layer BiLSTM) and self-attention), we design a series of model variants to further verify the effectiveness of each building module. In this section, we train MDER with or without each of those building modules in the ablation experiment. The details of the involved model variants, called MDER w/o rule, MDER w/o CNN and MDER w/o self-attention, respectively, are presented as follows: \par

\begin{itemize}
\item MDER w/o rule: The MDER model without the rule embedding module;
\item MDER w/o CNN: The MDER model without the CNN module;
\item MDER w/o self-attention: The MDER model without the self-attention module.
\item MDER w/o CRF: The MDER model without the CRF module.
\end{itemize}

\begin{table}[htbp]
    \centering
    \small
    \setlength{\tabcolsep}{4pt}
    \begin{tabular}{lccccccc}
    \toprule
    &Precision&Recall&F1 \\
    \midrule
    \multirow{1}{*}{w/o rule}  &0.7247&0.6157&0.6657 \\
    \cmidrule(lr){1-4}
    \multirow{1}{*}{w/o CNN}     &0.7157&0.6080&0.6574 \\
    \cmidrule(lr){1-4}
    \multirow{1}{*}{w/o self-attention}     &0.7047&0.5957&0.6455 \\
    \cmidrule(lr){1-4}		
    \multirow{1}{*}{w/o CRF}     &0.6721&0.5704&0.6171 \\
    \cmidrule(lr){1-4}
    \multirow{1}{*}{Complete MDER}     &$\textbf{0.7623}$&$\textbf{0.6420}$&$\textbf{0.6963}$ \\
    \bottomrule
    \end{tabular}
\caption{\label{Table: 4 Ablation result}  The effect of different module of our model on the mixed dataset.}
\end{table}

\begin{figure}
	\centering
		\includegraphics[scale=.46]{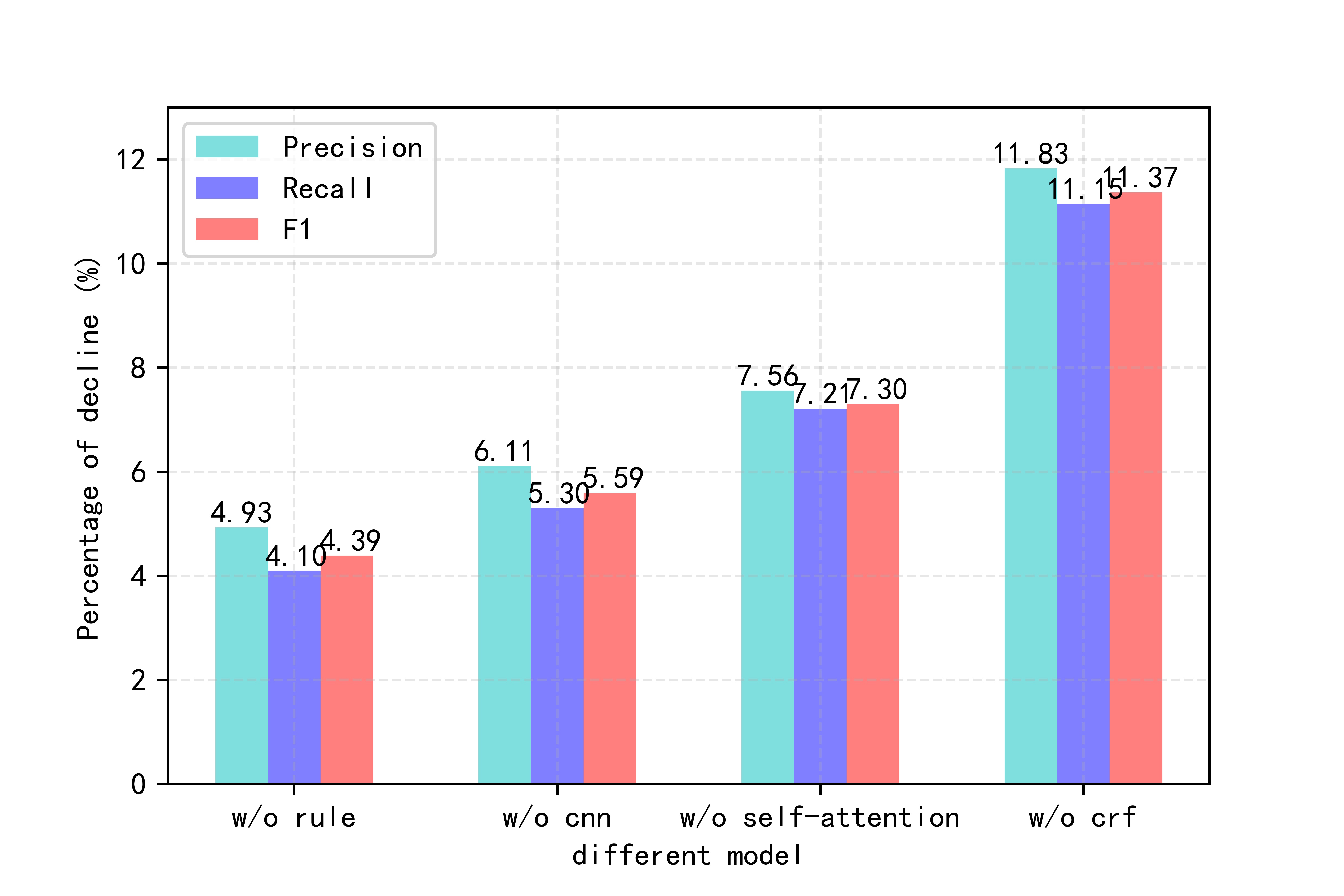}
	\caption{The performance drop of the three models relative to the original model on the mixed dataset.}
	\label{Figure: 4 histogram}
\end{figure}\par

We train and test all these model variants using the mixed dataset three times and compare them with the complete MDER model. Table \ref{Table: 4 Ablation result} shows the entity recognition performance of all these models. Unsurprisingly, all model variants with a certain key component being taken out achieve the inferior performance in terms of all the three performance measurements compared with the full MDER model. Figure \ref{Figure: 4 histogram} shows the performance drop (by percentage) of the four model variants compared to the full MDER model. It is easy to see from the histogram that the w/o CRF and self-attention models have the larger percentage of decline, which suggests that the two module have the greater impact on our model. The third is the w/o CNN model, while the w/o rule model has the slightest performance drop which shows that its influence on our model is the least, relative to the other model variants. \par
Despite their varying degree of importance, all the four key building modules are important to collectively contribute to the good performance of our model: rule embedding helps reduce the learning burden of the model and make the learning more efficient, CNN can capture the structural information according to the current context, which is a useful supplement to BiLSTM. The self-attention component focus on the important words related to the entity in the context bases on the interactive information between different words. CRF could better characterize the dependency between tags which is widely used.

\subsection{The Effect of Data Augmentation}

In this section, we will explore the impact of the different dataset sizes on the performance of MDER. Given the limited amount of real data available for all the four datasets, we resort to data augmentation to increase the data set size in a very efficient and economical way so that we can study how it will impact the performance of our model. To conduct data augmentation, based on the original 2,800 sentences in each dataset, we generate synthetic samples through entity substitution and add them into each dataset. For each dataset, we have one glossary for the method entities and another one for the dataset entities. Entities in sentences within the dataset are randomly replaced with other entities with the same type based on the corresponding glossary. \par

\begin{figure}[H]
	\centering
		\includegraphics[scale=.48]{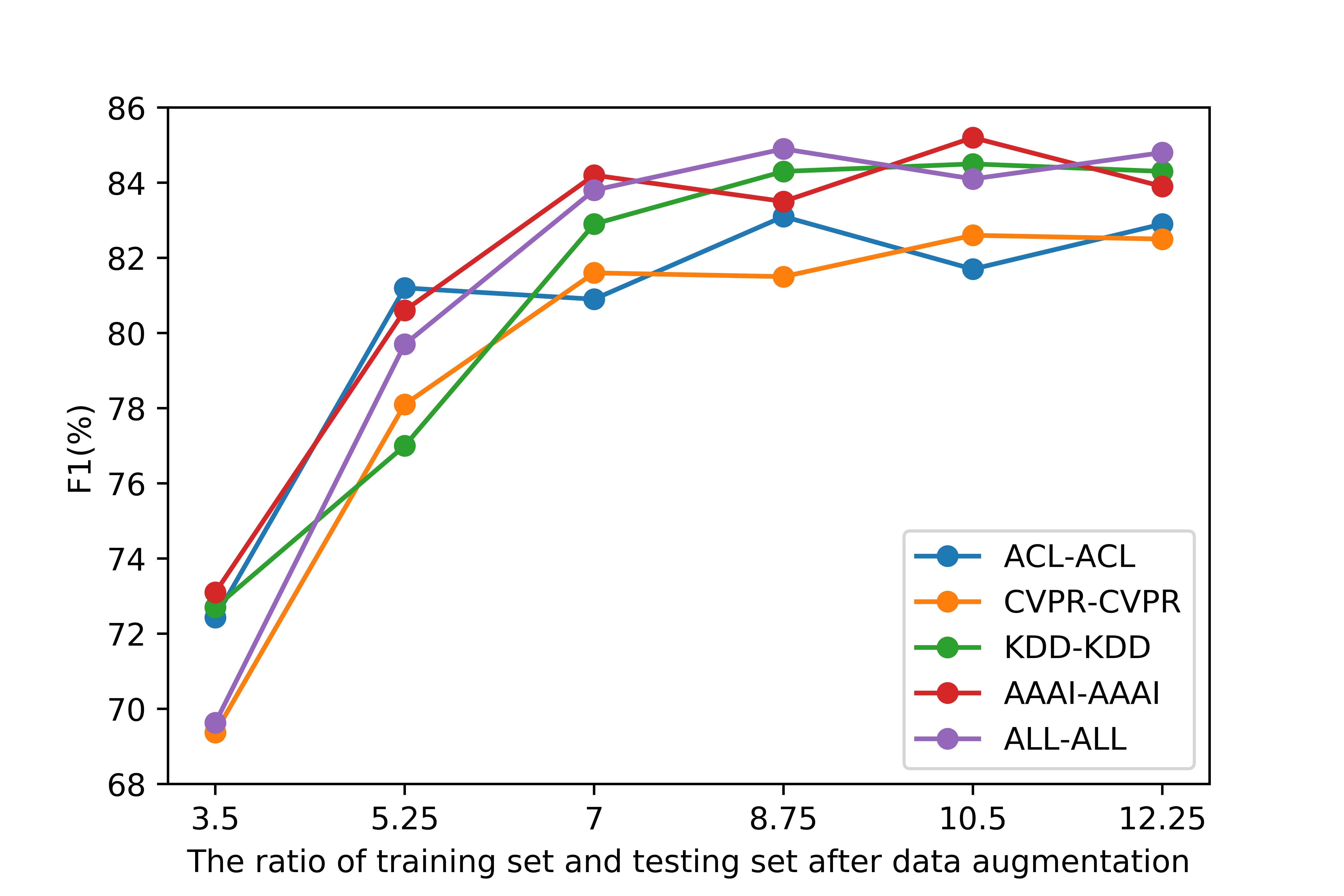}
	\caption{The result of our model on different datasets size.}
	\label{Figure: 5 augmentation}
\end{figure}

\begin{figure}[H]
	\centering
		\includegraphics[scale=.48]{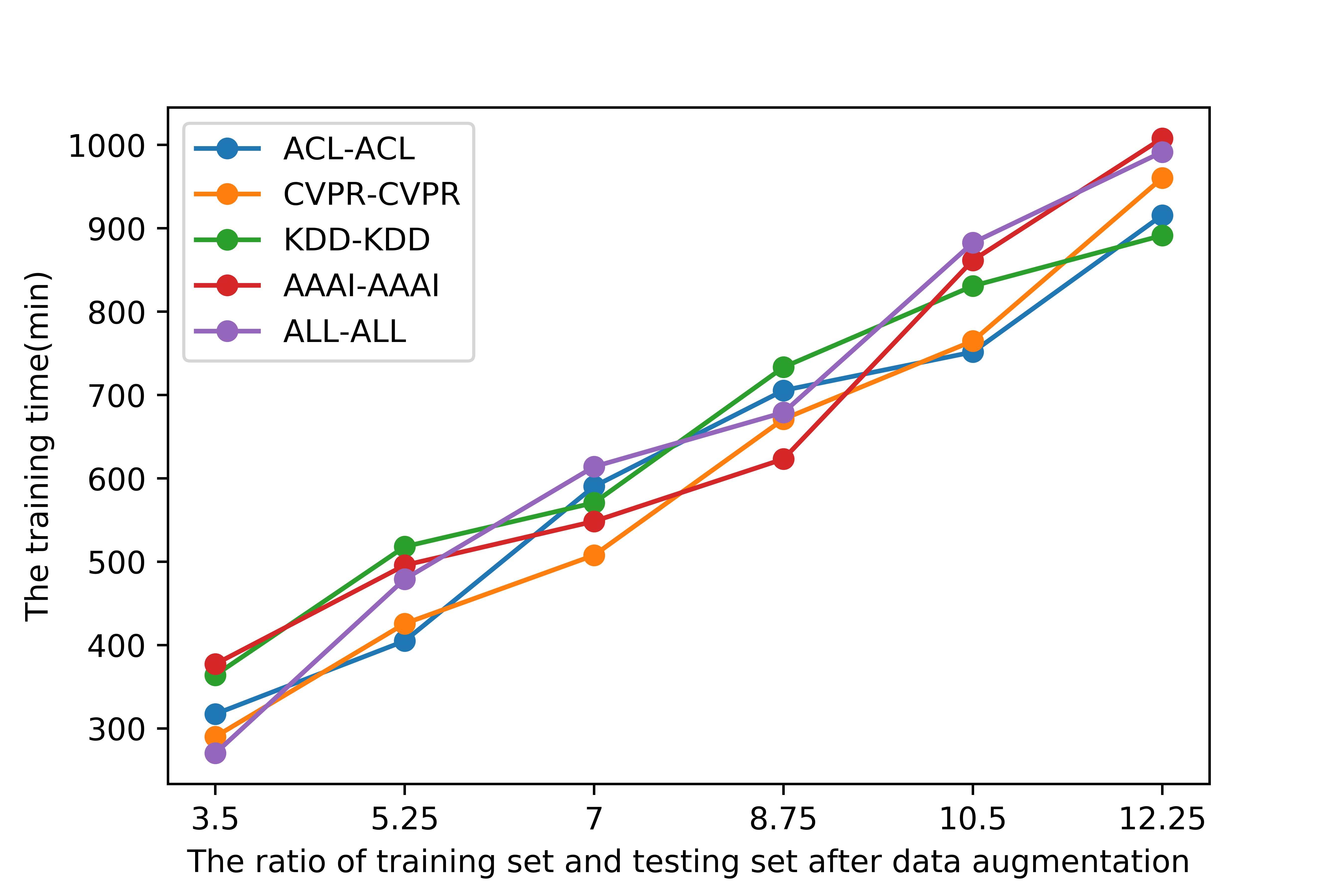}
	\caption{The result of the training time of the models with the increase of data.}
	\label{Figure: 6 training}
\end{figure}

Figure \ref{Figure: 5 augmentation} shows the variations of performances under different proportions of data augmentation on five datasets, and the five lines represent the F1 scores of the proposed model after augmentation is applied on the five datasets. The starting F1 value of each line correspondents to the diagonal value in Table 2 as both the training and test sets are from the same area. To maintain fairness and consistency, the model on each line is tested on the same testing set, i.e. the 20\% of the original 2,800 sentences, for each dataset. In the experiment, we take a 50\% increment for the amount of augmented data for each dataset. As a result, the size of the final dataset after augmentation is 1.5, 2, 2.5, 3, 3.5 and 4 times of the original dataset, respectively. From Section 4.2, we know the ratio of the training set to the test set is 7/2 = 3.5 on the original dataset (which is the starting value of the x axis of Figure \ref{Figure: 5 augmentation} and \ref{Figure: 6 training}). Hence, the final ratio of the size of the training set against the test set is 5.25, 7, 8.75, 10.25, 12.25, 14, respectively after data augmentation. For the augmentation of the mixed dataset, we still randomly sample from the four area-specific datasets as we do in Section 4.1. \par

\begin{figure}[H]
	\centering
		\includegraphics[scale=.48]{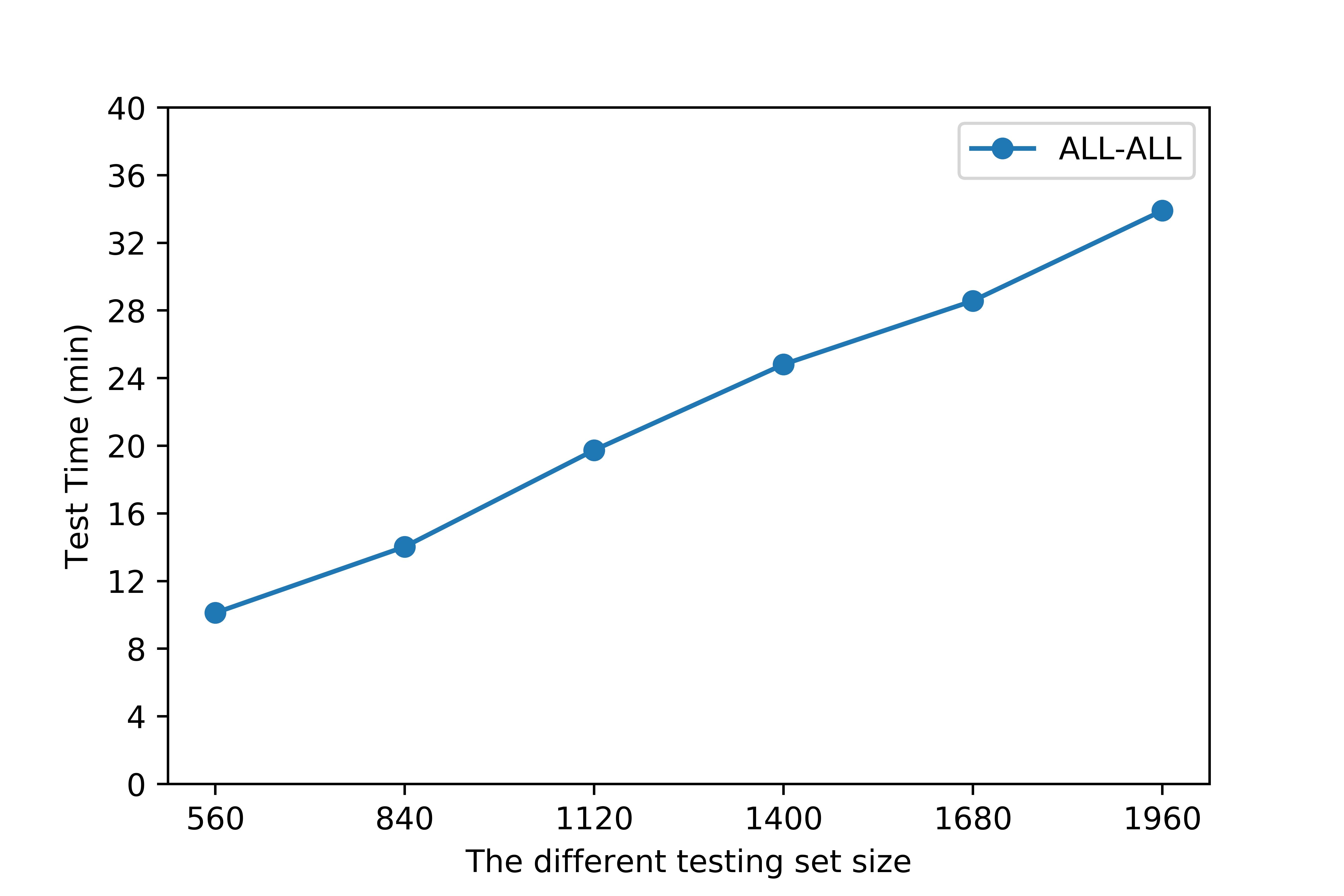}
	\caption{The result of the test time of the models with the increase of test data set.}
	\label{Figure: 7 test}
\end{figure}

As shown in Figure \ref{Figure: 5 augmentation}, the general trend of each line is same. When the number of training data increases at the beginning, the F1 value increases by a large margin. This is because that a larger training dataset can substantially increase the performance of our model at the beginning. However, after the ratio reaches beyond 7, the increasing trend for our model becomes gradually flattening even when more training data are added, meaning that the benefits of additional training data start to diminish. \par
Figure \ref{Figure: 6 training} and Figure \ref{Figure: 7 test} characterize the execution time of our model in the training and testing stages, respectively. Both figures show an approximately linear increase of the execution time of our model when the training and testing data sets increase in size, but the lines in Figure \ref{Figure: 6 training} feature a higher slope than that in Figure \ref{Figure: 7 test}, indicating that the training is more computationally expensive than the testing for our model. In other words, once our model has been trained, applying it to new papers for method and dataset mining is more efficient. Considering both the F1 values and the execution time presented in Figure \ref{Figure: 5 augmentation}, \ref{Figure: 6 training} and \ref{Figure: 7 test}, we can see that when the data ratio goes beyond 7, the linear increase of the training time does no longer justify the increasingly negligible gain in recognition performance.
\subsection{Long-term Literature Mining}

\begin{table}[htbp]
    \centering
    \small
    \setlength{\tabcolsep}{4pt}
    \begin{tabular}{lccccc}
    \toprule
    &ACL&CVPR&SIGKDD&AAAI&ALL \\
    \midrule
    \multirow{1}{*}{Precision}     &76\% &74\% &82\% &77\% &80\% \\
    \bottomrule
    \end{tabular}
\caption{\label{Table: 5 five models} The Precision results tested with five models.}
\end{table}

\begin{figure}[H]
	\centering
		\includegraphics[scale=.46]{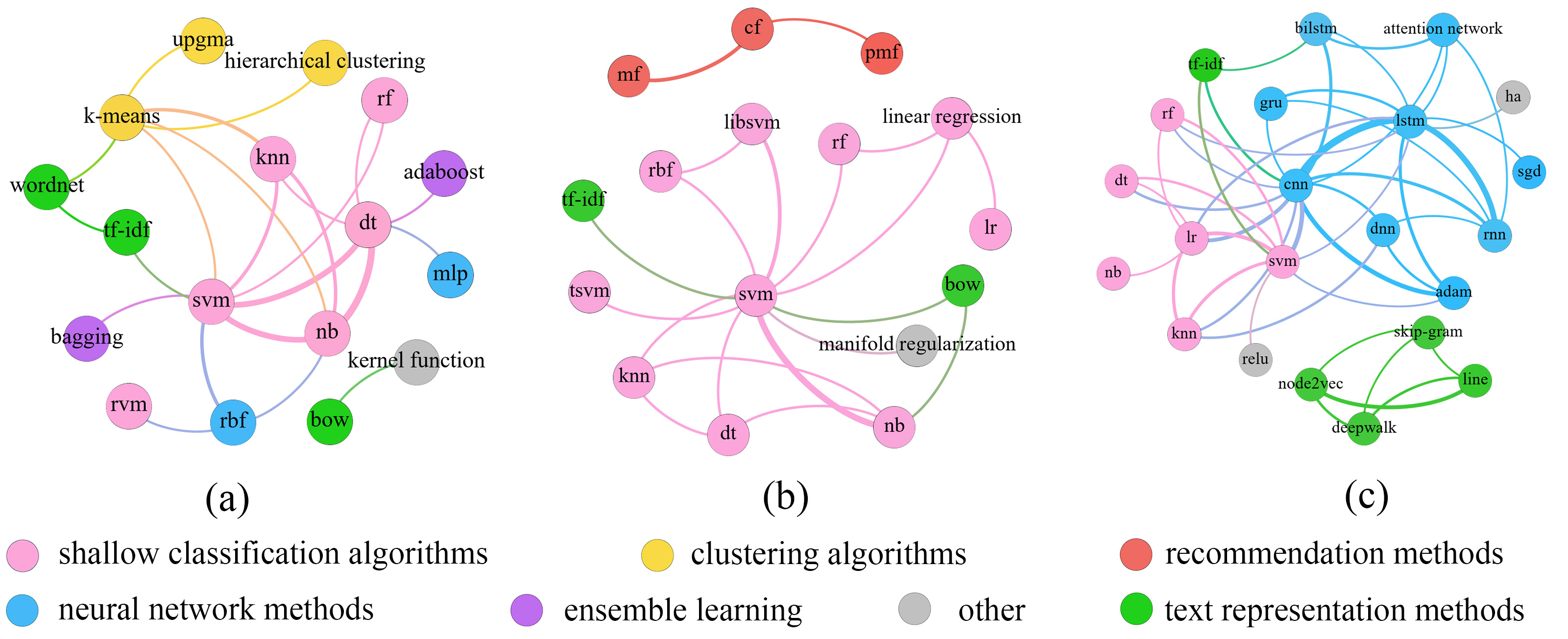}
	\caption{The method network with edge weights $>2$ in 2009, 2014 and 2019.}
	\label{Figure: 8 network}
\end{figure}\par

\begin{figure}
	\centering
		\includegraphics[scale=.23]{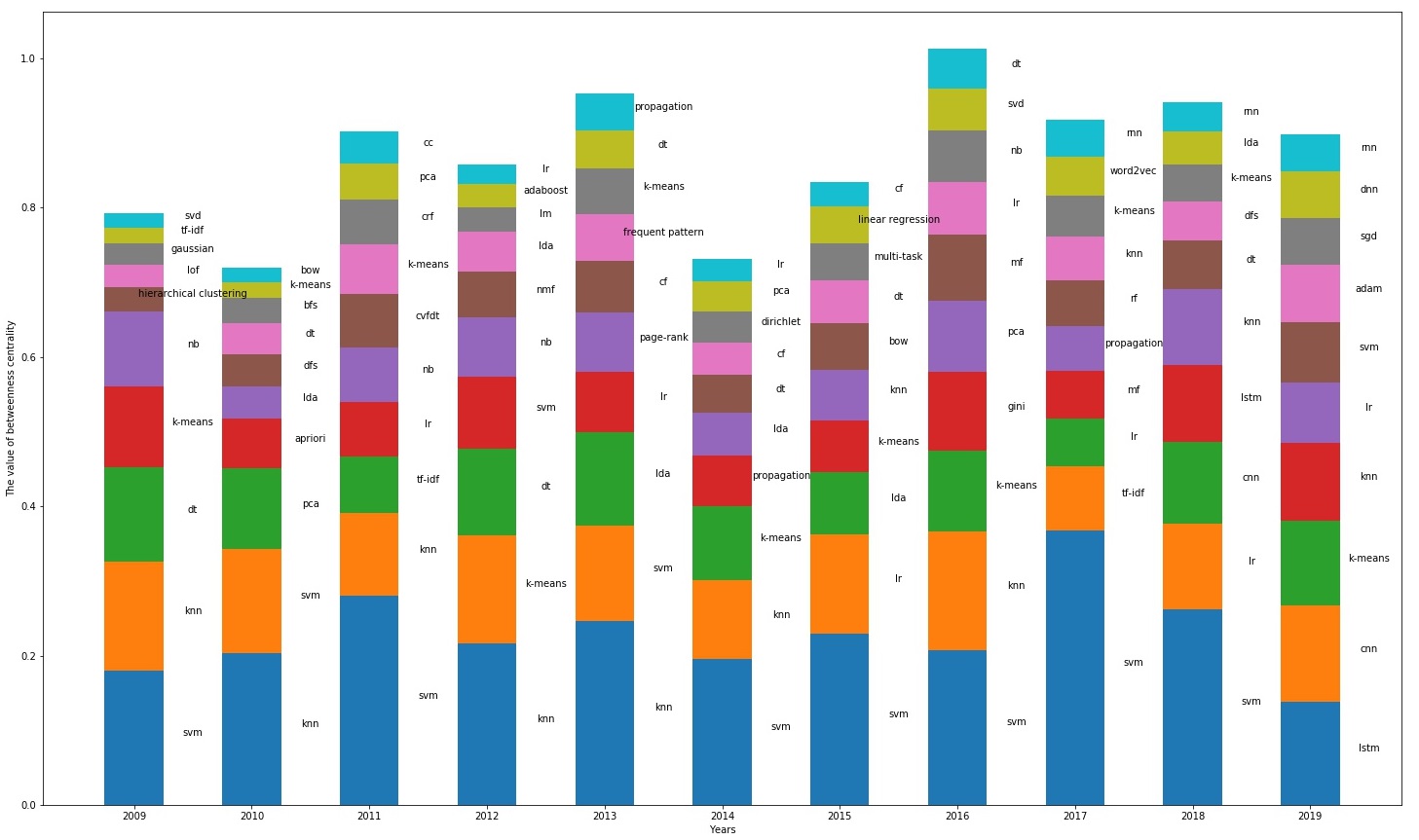}
	\caption{The histogram of the betweenness centrality of the top ten methods from 2009 to 2019.}
	\label{Figure: 9 betweenness}
\end{figure}

In this section, we analyze the long-term development (over 10 years) of the method and dataset entities involved in PAKDD publications to discover useful patterns and insights. The general idea is to train our model using the five different datasets (which leads to five different models) and apply them individually to extract the method and dataset entities appearing in PAKDD publications. The test dataset for this case study is composed of the sentences collected from the experimental section of 1,226 PAKDD conference papers published from 2009 to 2019. \par

Because the test data of PAKDD are unlabeled and it is very time-consuming to label all of them to obtain the ground truth result, it is thus difficult to accurately quantify the Precision, Recall, and F1-score of our model on the PAKDD dataset. As an approximate alternative evaluation approach, we take a sampling strategy to only evaluate the precision performance of each model. To be specific, we randomly select a specific number of predicted method and dataset entities (such as 50 each) from our model, and then manually validate the correctness of the entity tagging (which is very efficient thanks to the typically small number of recognized entities). If there are, for example, a total of 90 entities which have been labeled correctly by our model, then the precision is 90\%. Please note that we do not intentionally tune our models in order to achieve a high precision in this experiment. The models are trained in a normal way which tries to achieve a good balance between precision of recall in the first place. The precision results of the five models in this experiment are shown in Table \ref{Table: 5 five models}, which involves employing the sampling strategies multiple times and averaging the final Precision results. Again, because the entities in the PAKDD dataset are not labeled, the Recall as well as the F1 measure cannot be quantified even using the sampling strategy, so the Recall and F1 measure results are not reported here.\par

The result shows that the recognition performance of the model trained on the SIGKDD dataset is the best, which is quite easy to understand as both PAKDD and SIGKDD are conferences in the field of data mining. Therefore, we use the recognition results produced by the model trained on the SIGKDD dataset to further construct complex network graphs and the histogram of the betweenness centrality (Figure \ref{Figure: 9 betweenness}). These visualization results can better show the popular methods used in different areas of computer science in the 10-year period from 2009 to 2019 and how they evolved over time. \par

The complex network graphs are built for visualizing the extracted method entities every five years starting from 2009 (Figure \ref{Figure: 8 network}). An edge in the graphs represents that the two methods connected by the edge appear in the same paper, and the edge weight is the number of papers containing the two methods. To facilitate the drawing of the graphs, we manually delete the inconsistent entities in the result and unify the naming method entities in the form of lowercase abbreviation. Figure \ref{Figure: 8 network} (a)-(c) show the partial graph results whose edge weights are greater than 2. The color of the nodes, indicating the category of methods, is supplied through manually classification.\par

Figure \ref{Figure: 8 network} (a) shows that the shallow classification algorithms, such as support vector machine (SVM), decision tree (DT), $k$-nearest neighbor (KNN), naive Bayes (NB) and Random Forest (RF), often appeared together in many PAKDD papers in 2009. The clustering algorithms also received a lot of attention, and researchers prefer to use hierarchical clustering and k-means methods. Figure \ref{Figure: 8 network} (b) presents that the shallow classification algorithms are still widely used in 2014, such as SVM, DT, NB, Logistic Regression (LR), etc., as well as the variants of SVM such as LIBSVM and TSVM. Recommendation methods are also popular, such as Collaborative Filtering (CF) and Probabilistic Matrix Factorization (PMF). Figure \ref{Figure: 8 network} (c) demonstrates that the deep learning models dominate the landscape in 2019. For example, Convolutional Neural Network (CNN), Gated Recurrent Unit (GRU), Long Short-Term Memory (LSTM) and Recurrent Neural Network (RNN) co-occur usually. Some text representation methods such as skip-gram and node2vec become also popular. Meanwhile, the shallow machine learning models, such as SVM, DT, KNN and LR, are still being extensively used, even though not as dominating as a decade ago.\par

Figure \ref{Figure: 9 betweenness} shows the top ten methods based on the betweenness centrality analysis in the complex network graph for each year. The betweenness centrality indicates the importance of a node in the graph by counting the number of shortest paths between each pair of nodes in the graph that pass through this node. In a network, the larger this value is, the more important the node will be. From the figure, we can see that SVM has the largest betweenness centrality for seven years, indicating that it is closely related to other methods and have been commonly used as a baseline comparison model in many studies during this period. From 2009 to 2016, the shallow machine learning methods dominated the landscape, but deep learning became increasingly popular from 2017 and start to dominating the landscape in the area of data mining - LSTM rose from the fourth position in 2018 to the first one in 2019 followed by CNN in the second position. \par

We also analyze the patterns exhibited by the dataset entities extracted from the PAKDD papers.  Specifically, We study the number of papers using the same dataset entity. According to statistical observations, in 2009, the frequencies of machine learning datasets are high, such as UCI, Wine, Iris, 20newsgroup and WeBKB, etc. However, in 2019, deep learning datasets about text and images are popular, such as SemEval, Stanford CoreNLP, Twitter, MNIST, ILSVRC and YouTube Faces, etc. This is clearly consistent with the insights we obtained from the method mining regarding the general research development trajectory for computer science in the past decade.

\section{Conclusion and Future Research Directions}
In this paper, we study the problem of method and dataset mining in scientific papers via a semantic-based deep learning extraction model, called MDER. Our model, which combines the rule embedding technique and a CNN-BiLSTM-Attention-CRF structure, achieves competitive performance on extracting method and dataset entities in computer science papers. Through comprehensive experimental studies, we find that our model has great transfer learning ability and generalization performance on datasets of different domains, especially on the mixed dataset. Also, the ablation experiments indicate that each module of our model is complementary which together collectively contribute to the good recognition performance. By applying the trained MDER model on the published PAKDD papers published during 2009-2019, we can effectively mine and analyze the relationship among different methods and datasets and their development trajectory and trends over a long time span. Based on the success of this case study, we believe that our model is generic enough to be applied to mine method and dataset entities from other domains as well. \par
In the future, we will study additional interesting patterns around the recognized entities such as mining the opinion and sentiments of authors regarding different methods to achieve automatic performance assessment of different methods. We are also interested in building models that can automatically classify the entities to produce, for example, the categories of different methods, and offer personalized method and dataset recommendations. Finally, we plan to apply our model for mining method and dataset entities from literature in other appropriate domains, such as science and engineering, to further enhance its impact to the literature mining in those domains.

\printbibliography{}
\typeout{get arXiv to do 4 passes: Label(s) may have changed. Rerun}
\end{document}